\documentclass[letterpaper, 10 pt, conference]{ieeeconf}

\IEEEoverridecommandlockouts
\usepackage{graphics}
\usepackage{epsfig}
\usepackage{mathptmx}
\usepackage{times}
\usepackage{amsmath}
\usepackage{amssymb}
\usepackage{url}
\usepackage{cite}
\usepackage{booktabs}
\usepackage{tikz}
\usepackage{graphicx}
\usepackage{pgf}
\usepackage{multirow}
\usepackage{array}
\usepackage{tabularx}

\usetikzlibrary{automata, positioning, arrows.meta, shapes.geometric, calc}
\usepackage{siunitx}
\usepackage{hyperref} 

\title{\LARGE \bf
Characterization, Analytical Planning, and Hybrid Force Control for the
Inspire RH56DFX Hand
}

\author{Xuan Tan, William Xie, and Nikolaus Correll%
\thanks{Authors are with the Department of Computer Science, University of Colorado
        Boulder. Corresponding email: {\tt\small \{xuan.tan, william.xie\}@colorado.edu}}%
}
\begin{document}

\maketitle
\thispagestyle{empty}
\pagestyle{empty}

%%%%%%%%%%%%%%%%%%%%%%%%%%%%%%%%%%%%%%%%%%%%%%%%%%%%%%%%%%%%%%%%%%%%%%%%%%%%%%%%
\begin{abstract}
Commercially accessible dexterous robot hands are increasingly prevalent, but many remain difficult to use as scientific instruments. For example, the Inspire RH56DFX hand exposes only uncalibrated proprioceptive information and shows unreliable contact behavior at high speed (up to $+1618\%$ force limit overshoot). Furthermore, its underactuated, coupled finger linkages make antipodal grasps non-trivial.
We contribute three improvements to the Inspire RH56DFX to transform it from a black-box device to a research tool: (1) hardware characterization (force calibration, latency, and overshoot), (2) a sim2real validated MuJoCo model for analytical width-to-grasp planning, and (3) a hybrid, closed-loop speed-force grasp controller.
We validate these components on peg-in-hole insertion, achieving 65\% success and outperforming a wrist-force-only baseline of 10\% and on 300 grasps across 15 physically diverse objects, achieving 87\% success and outperforming plan-free grasps and learned grasps. 
Our approach is modular, designed for compatibility with external object detectors and vision-language models for width \& force estimation and high-level planning, and provides an interpretable and immediately deployable interface for dexterous manipulation with the Inspire RH56DFX hand, \href{https://correlllab.github.io/rh56dfx.html}{open-sourced at this website}.
\end{abstract}

%%%%%%%%%%%%%%%%%%%%%%%%%%%%%%%%%%%%%%%%%%%%%%%%%%%%%%%%%%%%%%%%%%%%%%%%%%%%%%%%
\section{INTRODUCTION}
\vspace{-1mm}
Commercial dexterous hands offer cost and accessibility advantages over research platforms, but often lack the characterization and control methodology needed for reliable precision manipulation using classical methods.
The Inspire RH56DFX exemplifies this gap: it provides six actuated DOFs and capable actuators, yet exposes raw, unitless signals for force (0 to 1000) and exhibits severe contact overshoot at high speed, limiting its utility as a research instrument.
\begin{figure}[t]
  \centering
  \includegraphics[width=0.99\columnwidth]{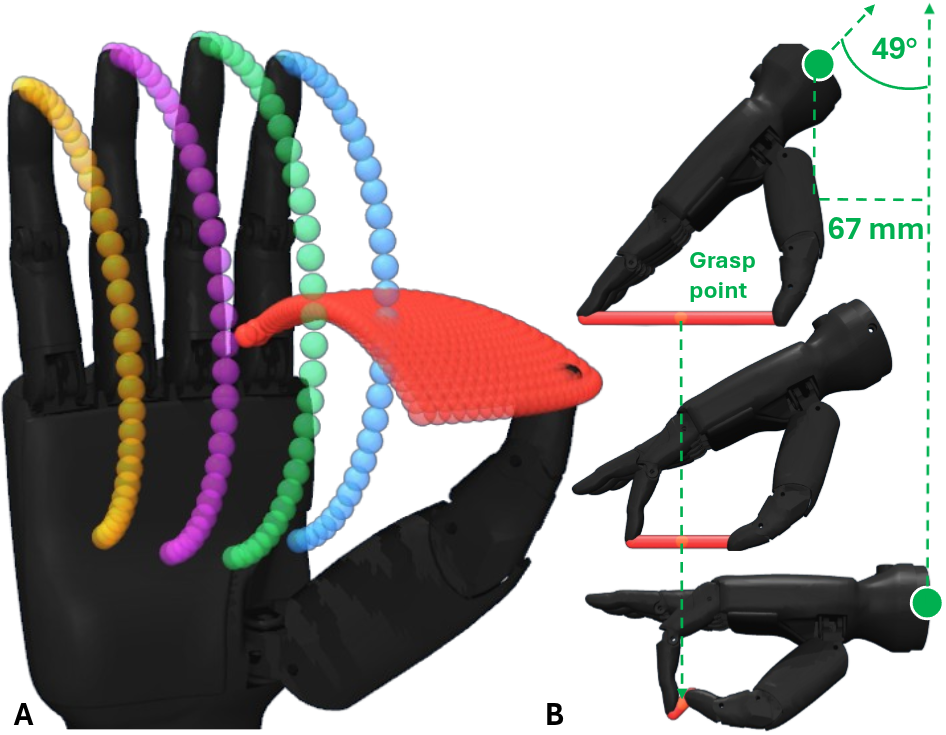}
  \caption{\textbf{A.} 3D workspace visualization of the Inspire RH56 hand, showing that the coupled finger linkages constrain motion to arcs. The red bubbles represent the workspace of the 2-DoF thumb yaw and pitch actuation, highlighting the limited region of intersection with the other fingers and subsequent constrained grasping. \textbf{B.} The coupled kinematics require significant translation and rotation up to $\ang{49}$ for different width, here 110, 55, and 0mm are shown. 
  \vspace{-2mm}
  \vspace{-1mm}
  %
  %For three two-finger pinch grasps for object widths 110, 55, and 0mm, the hand must tilt $\ang{49}$, and, in addition, adjust 7 mm downward and 12 mm laterally, not depicted, to maintain a grasp about a point in space.
  }
  \label{fig:workspace_2d}
\end{figure}
In addition, coupled linkages make the resulting kinematics non-trivial, requiring different transformations (rotation and translations) as a function of object width (Figure \ref{fig:workspace_2d}). While imitation learning approaches \cite{FromPowerToPrecision2024} can ignore these intricacies, applying classical methods \cite{rimon2019mechanics} requires significant characterization, system identification, and planning. 

This paper demonstrates that a commercially available hand can be transformed into a research-grade platform through rigorous characterization, analytical modeling, and physics-aware control---without requiring learning or hardware modification.

%%% I THINK THIS PARAGRAPH ANTICIPATES THE RESULTS
%For under-actuated commercial dexterous hands, explicit system characterization combined with analytical geometry and low-speed force-aware execution yields reliable grasping comparable to dedicated research platforms.
%By investing in software-level characterization instead of hardware replacement, we convert raw load units to calibrated force (R$^2 > 0.98$), expose the coupled-linkage geometry that shifts fingertip height by 75\,mm during closure, and derive a width-parameterized antipodal planner.
%The resulting control stack prioritizes interpretability and reproducibility: pre-position quickly in free space, close slowly during contact, and regulate force with intrinsic finger-mounted sensors rather than wrist F/T estimates.

%%%%% I THINK THIS PARAGRAPH ANTICIPATES TOO MUCH OF THE RESULTS, THE INTRODUCTION SHOULD JUST MOTIVATE THE WORK
%On a vision-free peg-in-hole benchmark, this approach achieves 13/20 success (65\%) using intrinsic-force release, substantially outperforming a wrist-force-only baseline (2/20, 10\%).
%The force-profile comparison explains the gap: wrist-threshold release exhibits substantially higher contact forces and higher variance than finger-threshold release, consistent with its lower success rate.
%This capability uplift on affordable hardware demonstrates the potential of physics-grounded characterization to narrow the gap between commercial and research-grade systems.

While learned end-to-end methods for dexterous grasping have advanced rapidly, they remain difficult to deploy on new hardware without retraining, and their failure modes are difficult to diagnose.
Our contribution is orthogonal: a fully interpretable, data-free pipeline that can be fully simulated and validated before deployment, composes with any perception module without modification, and provides unique per-finger force and force-closure diagnostics that black-box learned policies cannot.
We do not claim to surpass state-of-the-art learned methods in raw success rate; we demonstrate that a principled classical approach---rigorously characterized and geometrically grounded---is competitive on real hardware and uniquely interpretable. We contribute:
\begin{itemize}
  \item \textbf{Hardware Characterization:} Force calibration (Shimpo gauge, R$^2 > 0.98$), latency quantification (66\,ms), and overshoot characterization

  \item \textbf{Hybrid Force Control:} Velocity-switching policy (fast in free-space, slow at contact) with intrinsic finger-force release thresholds, demonstrated in peg-in-hole insertion with 65\% success rate.

  \item \textbf{Analytical Grasp Planning:} MuJoCo model with width-parameterized antipodal grasp planner with online force-closure visualization, validated by 300 grasps across 15 diverse objects, achieving 87\% success.

\end{itemize}
\subsection{Related Work and Positioning}
%\subsubsection{Competing Dexterous Hand Platforms}
Research-scale dexterous hands span a spectrum from fully-custom research prototypes to increasingly capable commercial platforms.
Higher-end research hands (DLR/HIT II, DEXMART, Shadow Dexterous Hand) offer dense tactile sensing, direct joint torque feedback, and sophisticated real-time control stacks, enabling advanced impedance and compliance strategies~\cite{Chen2011_DLR_Impedance}.
These platforms excel at contact-rich tasks but typically carry \$$50$k--$200$k acquisition costs and limited widespread availability.
The Inspire RH56DFX and its successor RH56FTP, which shares identical kinematic constraints, occupy a complementary niche: it is commercially available, affordable (approximately \$8000), and adopted across industry and research labs.
Our work demonstrates that this cost accessibility can be leveraged by investing in software-level characterization and control rather than requiring hardware replacement.
%The resulting system is not a substitute for research hands with integrated torque sensing, but shows that data-free analytical methods can narrow the gap in practical task reliability (65\% peg-in-hole vs.~10\% baseline), making commercial hands viable for interpretable research pipelines.

%\subsubsection{RH56DFX Applications and Control}

RH56-focused literature is growing, but most works use the hand as an integrated end-effector rather than as a characterized control substrate.
In dynamic grasping, Han \emph{et al.} pair an AUBO arm with RH56DFX-2R and report strong moving-target pickup, but without a detailed intrinsic-force control analysis or open-source release~\cite{Han2025Tracking}.
In teleoperation and imitation, RH56 is used to replay human intent (markerless vision + sEMG, glove-driven imitation, and cross-task policy learning), with emphasis on high-level policy transfer and limited reporting of low-level contact regulation or RH56-specific calibration~\cite{Liu2025Teleop,Sharma2025Assistive,OminiAdapt2025}.
In perception/sensor integration, tactile and flexible sensors are mounted on RH56 platforms for recognition and performance evaluation; these studies validate sensing quality but typically rely on position-driven execution and do not provide a complete RH56 force-control stack~\cite{Lu2022Tactile,Yu2024Skin,Yin2025Single,Liang2025Nanocrack}.
Affordance-driven task planning has also been demonstrated on robots equipped with RH56, but low-level dexterous execution is usually abstracted away~\cite{Glover2024}.

Our contribution complements these learning- and sensing-centric pipelines with a hand-centric, reproducible RH56 methodology: explicit force calibration and dynamic characterization, analytical width-to-grasp planning from validated kinematics, and a hybrid intrinsic-force policy that enforces low-speed contact where RH56 dynamics are stable.

Recent surveys show that dexterous manipulation performance depends strongly on contact modeling fidelity and force-aware execution, whether in model-based planning or RL pipelines~\cite{Yu2022DexterousReview,Weinberg2024InHandSurvey,Song2025DexterousOverview,Huang2025HumanlikeReview}. Contact-model analyses further show that numerical formulation choices can materially affect contact realism and optimization behavior~\cite{LeLidec2024ContactModels,Jin2024ComplementarityFree}, motivating careful simulator-grounded validation.

On the planning side, two-stage and taxonomy-level grasp synthesis methods scale grasp generation and contact feasibility across hand-object pairs~\cite{Tian2023TwoStage,Dexonomy2025, 1241860, 1371616}, while simulation studies emphasize slip/stability prediction in MuJoCo-like engines. Our work is complementary: we use MuJoCo for analytical grasp synthesis and force-aware execution design, then validate on RH56 hardware with calibrated intrinsic sensing, enabling real-hand force trends that match simulation structure without demonstration-scale learning.

%%%%%%%%%%%%%%%%%%%%%%%%%%%%%%%%%%%%%%%%%%%%%%%%%%%%%%%%%%%%%%%%%%%%%%%%%%%%%%%%
\section{SYSTEM CHARACTERIZATION}
\label{sec:characterization}

To transform the RH56 into a research instrument, we first characterize force sensing and contact dynamics.

%\subsection{Intrinsic Force Calibration}
%
%Calibration used a Shimpo FGV-10XY force gauge.
To perform force calibration, we let each finger push directly against a force gauge (Shimpo FGV-10XY) while the internal RH56 force parameter is swept from 25 to 1000 in increments of 25, with manual gauge readings recording at each step.
Every point is collected twice to verify repeatability and consistent sensor output.

Raw load feedback $L_\text{raw}\in[0,1000]$ is converted to Newtons with a per-finger linear model:
\begin{equation}
F_\text{est}=aL_\text{raw}+b.
\end{equation}

Across index, middle, and thumb bend actuators, the fit is highly linear ($R^2>0.98$), enabling metric force feedback.
\begin{table}[t]
\tiny
\caption{Intrinsic force calibration coefficients ($F=aL_\text{raw}+b$).}
\vspace{-2mm}
\label{tab:force_coeffs}
\centering
\footnotesize
\begin{tabular}{lccc}
\toprule
Finger & Slope $a$ & Intercept $b$ & $R^2$ \\
\midrule
Index  & 0.0075 & $-$0.414 & 0.987 \\
Middle & 0.0065 & 0.018    & 0.986 \\
Thumb bend & 0.0125 & 0.384 & 0.993 \\
\bottomrule
\end{tabular}
\vspace{-1mm}
\vspace{-1mm}
\vspace{-1mm}
\vspace{-1mm}
\end{table}
\begin{figure*}[!htb]
  \centering
  \resizebox{0.9\columnwidth}{!}{\input{plot/figure_step_response.pgf}}
  \includegraphics[width=\columnwidth]{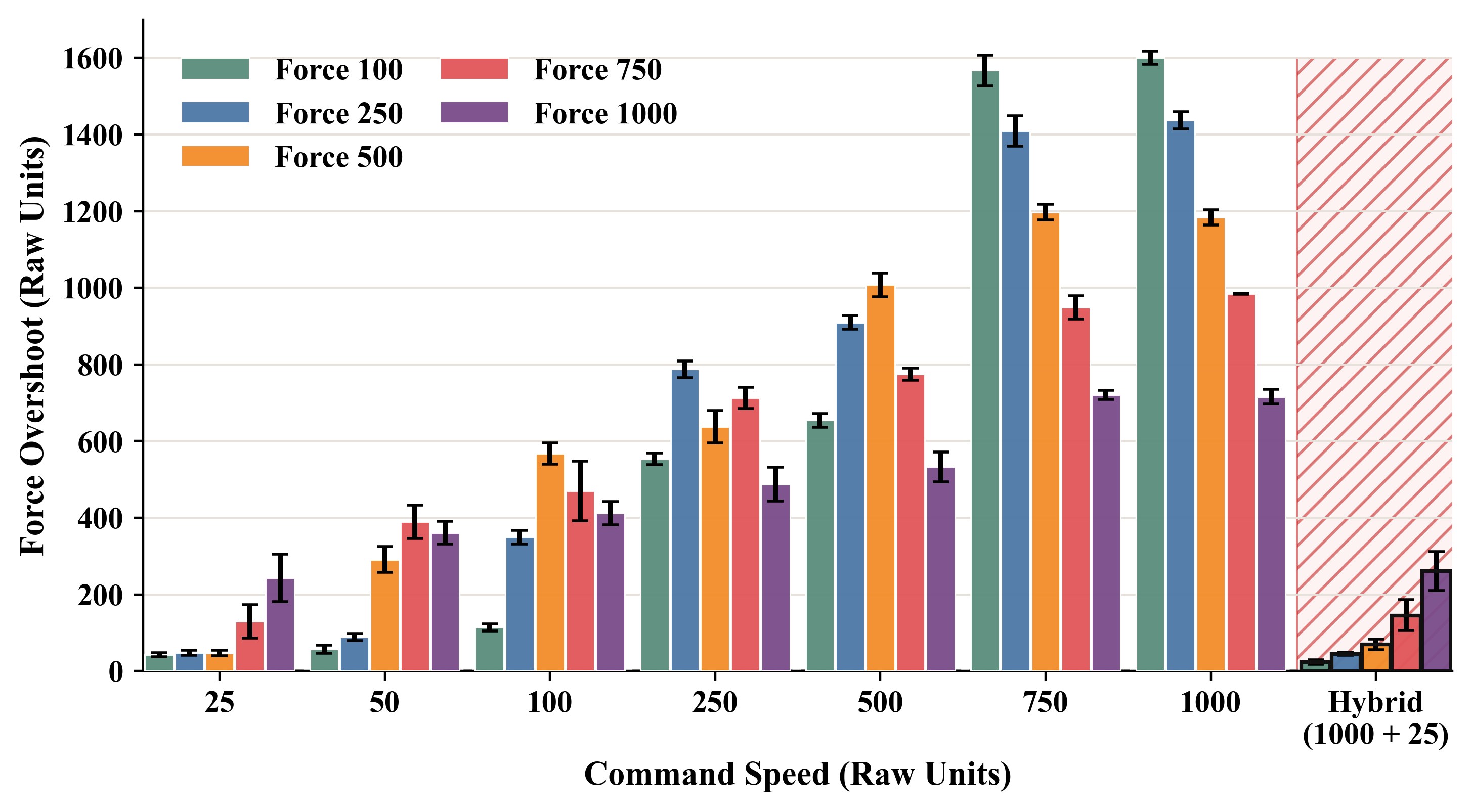}
  \vspace{-1mm}
  \vspace{-1mm}
    \caption{\textbf{Left:} Step response to reach position ``500'' for different speed settings, showing  66ms latency followed by linear growth. The fingers stop when the set-points are reached with significant overshoot in the force domain (see below).
  \label{fig:step_response}
  \textbf{Right:} Force overshoot as a function of commanded speed and force setpoint ($N=20$, error bars = variance).
  Overshoot increases with higher speed, while hybrid control yields overshoot close to that of constant speed 25.
  \label{fig:overshoot}}
  \vspace{-1mm}
  \vspace{-1mm}
  \vspace{-1mm}
  \vspace{-1mm}
\end{figure*}

\begin{figure}[!htb]
  \centering
  \resizebox{0.8\columnwidth}{!}{\input{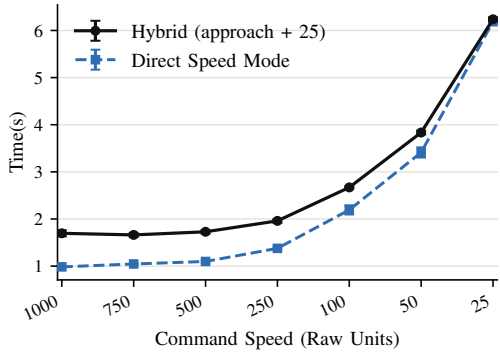}}
  \vspace{-1mm}
  \vspace{-1mm}
  \vspace{-1mm}
  \caption{Completion time vs. commanded speed under identical force setpoint for various speeds and two different control parameters. %We compare two velocity control schemes across commanded speeds 1000 to 25 (raw units), with 20 trials per setting. 
  In Direct Speed Mode, a single constant speed is commanded for the entire motion. In the Hybrid scheme, %the approach speed varies (1000 to 25) while 
  the contact speed is fixed at 25. 
  %Time is measured from the first command transmission to the moment the maximum force is reached. 
  Markers show the mean over trials ($N=20$) and error bars indicate variance. %The time gap narrows as the approach speed decreases, and the curves coincide at 25 as a sanity check since both schemes effectively operate at 25 throughout.
  }
  \label{fig:completion_time}
  \vspace{-1mm}
  \vspace{-1mm}
  \vspace{-1mm}
  \vspace{-1mm}
  \vspace{-1mm}
\end{figure}
%\subsection{Dynamic Constraints for Contact}
To characterize the actuator's dynamic response, we performed step-response tests: commanding the index finger to move at velocities ranging from 100 to 1000 toward a target position of 500 (raw encoder units). Sensors are read and commanded at 163\,Hz.

Fig.~\ref{fig:step_response}, left, reveals that after an initial 66\,ms latency (command transmission to first sensor reading), the finger position grows \textit{linearly} without deceleration, indicating no active speed ramping or position feedback correction.
This linear motion persists through the response window, leading to rise times of 0.18--0.30\,s and settling times of 0.27--0.43\,s.
The absence of deceleration near the target has severe consequences for force control: the finger does not slow down as it approaches contact, resulting in high-speed impact when the object is encountered.

The latency-plus-linear-motion behavior cascades into the force domain.
When a finger traveling at high speed makes first contact with an object, the 66\,ms latency prevents immediate velocity reduction.
During this latency, the finger continues at full speed, compressing the object and generating excessive force---an effect we quantify next.

For contact characterization, we repeatedly drive a single finger into a rigid cube (4\,cm edge length) while varying both the commanded speed and the force setpoint.
We evaluate constant-speed commands $v\in\{25,50,100,250,500,750,1000\}$ and, motivated by modern, capable 3D object detection methods for object segmentation and width estimation ~\cite{DeliGrasp}, evaluate a hybrid profile that uses a fast approach ($v=1000$) to a specified position followed by a low contact speed ($v=25$).
For each speed condition and each force setpoint $F_{\text{set}}\in\{100,250,500,750,1000\}$ (raw units), we ran 20 trials and computed the overshoot
$\Delta F = F_{\max}-F_{\text{set}}$ within each trial.
Fig.~\ref{fig:overshoot}, right, reports the mean overshoot with error bars indicating trial-to-trial variability.

Overshoot increases sharply with commanded speed across all setpoints, whereas low-speed contact ($v=25$) yields consistently small overshoot; thus, the hybrid profile produces overshoot close to constant $v=25$ (Fig. \ref{fig:completion_time}).
Given the measured command-to-sensing latency (approximately 66\,ms), Fig.~\ref{fig:step_response}, left, further shows that the finger advances at an approximately constant velocity with no observable pre-deceleration before the target position.
As a result, the commanded stop (or force-limit transition) is applied only after the finger has already traveled an additional distance during the latency window, making high-speed impacts unavoidable and directly amplifying force overshoot.
These results motivate the hybrid execution: move quickly in free space to an offset position, then switch to low-speed contact mode ($v\leq25$) near contact, bounding overshoot while maintaining fast overall execution.

%%%%%%%%%%%%%%%%%%%%%%%%%%%%%%%%%%%%%%%%%%%%%%%%%%%%%%%%%%%%%%%%%%%%%%%%%%%%%%%%
\vspace{-1mm}
\subsection{Hybrid Force Control Policy}
\vspace{-1mm}
Let $F_\text{est}$ be calibrated intrinsic force and $F^\star$ the target.
Execution is bimodal: (1) fast pre-position in free space and (2) slow contact approach at $v\leq 25$.
% Execution is bimodal: (1) fast pre-position in free space and (2) slow contact approach at $v\leq 25$, and (3) deadband hold around $F^\star$.

For each closure move, let $q_0$ be the initial command position and $q_g$ the goal position in RH56 command units. We switch from fast to slow mode at a 25 distance threshold:

\begin{equation}
% q_\text{sw} = q_g + \alpha (q_0-q_g),\qquad \alpha=0.1.
q_\text{sw} = q_g + 25.
\end{equation}

We choose the fixed 25-unit margin empirically from the repeatability of the detected contact onset.
Across 800 trials, the largest variability occurs at the highest speed ($v=1000$), with a standard deviation of approximately $\sigma_{\text{onset}}\approx 7.5$ command units.
Setting $q_{\text{sw}} = q_g + 25$ corresponds to a conservative buffer of about $25/7.5 \approx 3.3\sigma$, which would cover $>99\%$ of onset variation under a near-Gaussian assumption and thus ensures the controller reliably enters low-speed mode before reaching the uncertain onset region.

% In slow mode, force is regulated with a deadband around $F^\star$:
% \begin{equation}
% q_{k+1}=\begin{cases}
% q_k+s\,u_c, & F_\text{est}<F^\star-\Delta F,\\
% q_k-s\,u_o, & F_\text{est}>F^\star+\Delta F,\\
% q_k, & |F_\text{est}-F^\star|\leq\Delta F,
% \end{cases}
% \end{equation}

% where $s=\operatorname{sign}(q_g-q_0)$ is the closing direction, and $u_c,u_o>0$ are small command steps.

% We additionally use a reflex closure sequence (see \ref{fig:grasp_strategies}): thumb first provides a geometric backstop, then opposing fingers close under this force policy. This reduces object ejection and improves repeatability for small objects.

\section{Analytical Grasp Planning with MuJoCo}
We design a MuJoCo-simulated RH56 model with calibrated joint limits and coupling constraints. We leverage the trajectory data collected in hardware characterization to do system identification in MuJoCo with least-squares optimization to find corresponding feedforward, damping, armature, and joint friction simulation values for the evaluated speed limit profiles in the previous section. Primarily, we leverage two configurations at the maximum and minimum speed limits (1000, 25) for usage in hybrid speed-force control.
Then, for generating antipodal grasps (a longstanding grasp synthesis technique of selecting two opposing, collinear contact points and surface normals on an object, typically perpendicular to its major axis, for stable, force-closure grasps ~\cite{FerrariCanny1992}), the central geometric challenge is a non-trivial closure-dependent tilt, up to $\ang{49}$ from full opening to closure, and fingertip displacement (7\,mm up and 12\,mm laterally), causing naive ``close-to-width'' commands to be systematically misaligned.

%\subsection{Width-to-Grasp Analytical Mapping}

We precompute fingertip forward kinematics offline by sweeping a MuJoCo model of the hand.
Let $\mathbf{d}=\mathbf{T}-\mathbf{C}$ be the thumb ($\mathbf{T}$)-to-reference-finger ($\mathbf{C}$) vector in the hand frame.

Coplanarity yields $\theta^\star = \operatorname{atan2}(-d_z,\,d_x)$, and the desired pinch width is $W = \sqrt{d_x^2+d_z^2}.$

Rather than searching independently over closure and tilt---which produces discontinuous solutions across the width range---we parameterize the closure state by a single scalar $s\in[0,1]$, where $\mathrm{ctrl}(s)=\mathrm{ctrl}_{\min}+s(\mathrm{ctrl}_{\max}-\mathrm{ctrl}_{\min})$, and $\mathrm{ctrl}$ are the control limits of the joint actuator.
The XZ-distance $D(s)=\sqrt{d_x^2+d_z^2}$ is monotone in $s$; Brent's scalar root-finding method~\cite{Birglen2008} locates $s^*$ such that $D(s^*)=W$ in sub-millisecond time.
The coplanarity tilt $\theta^*$ then follows from $s^*$ as above, producing a smooth, offline solution across all reachable widths with guaranteed convergence. 

\begin{table}[h]
\scriptsize
\caption{Reachable widths (\textit{mm}) and quality metrics by \textit{n}-finger pinch.}
\label{tab:width_ranges}
\centering
\renewcommand{\arraystretch}{1.08}
\begin{tabular}{@{}lrrrrrr@{}}
  \toprule
Fingers & Min & Max & Z-span & Z-span, \textit{QP} & Tip err. \textit{QP} \\
\midrule
2 & 0  & 110 &  1.2 &  2.9 & 2.4 \\
3 & 0  & 100 &  7.7 &  6.9 & 2.6 \\
4 & 0  & 100 &  8.2 &  6.9 & 2.7 \\
5 & 0  & 100 & 21.6 & 18.3 & 3.3 \\
\bottomrule
\end{tabular}
\end{table}

%\textit{Multi-finger extension.}
For plane grasps with $n\in\{3,4,5\}$ fingers, all non-thumb fingers share the same $s^*$ from the reference (index) finger solve, so all fingertips reach coplanar contact simultaneously.
The thumb is then solved independently to span the target width from the opposing side.
The residual Z-span (Table~\ref{tab:width_ranges}) measures the average Z-axis deviation from the grasp plane for each finger and reflects physical geometry: below 2\,mm for 2-finger grasps, growing moderately with $n$ due to the physical spread of finger bases along the palm. For the two-finger pinch, we obtain a (min, max) grasping range of (0, 110) mm for the 2-finger pinch grasp and (7, 100) mm for the 3, 4, and 5-finger pinch grasps. This analysis gives a direct, interpretable mapping from measured object width to full hand configuration. Then, given a desired $d_x, d_z, \theta$ for the hand's fingers, a known transformation from the fingers to a fixed hand base to a high-DoF robotic manipulator, and an inverse kinematics solver, the hand can be sufficiently controlled to width-parameterized grasp pose. 

\begin{figure}[!htb]
  \centering
  \includegraphics[width=0.95\columnwidth]{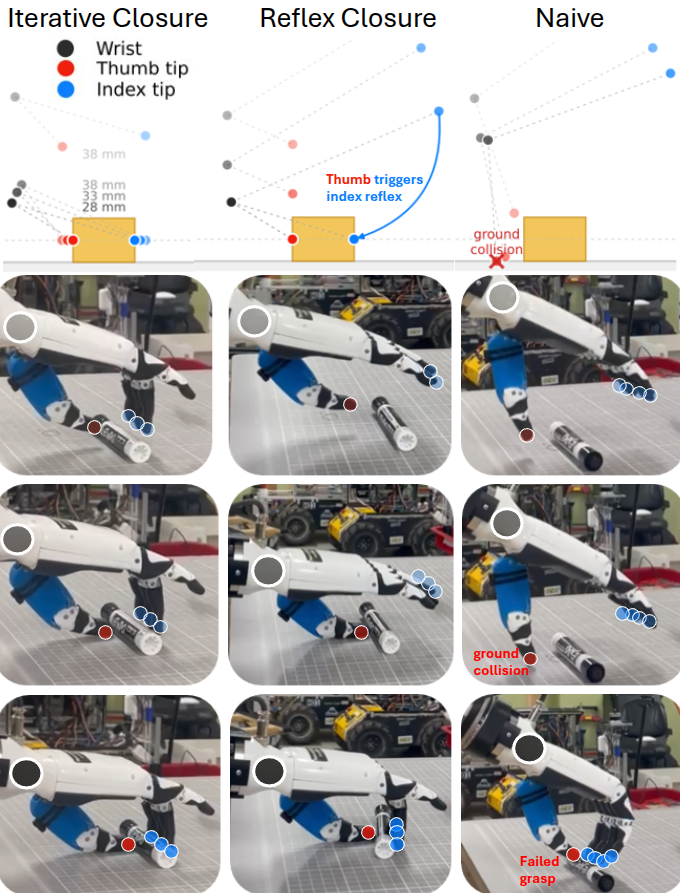}
  \caption{We formulate three grasp strategies: iterative (left), reflex (center), and naive (right) closure. In the iterative approach, the hand rotates downward from a pre-grasp pose to reach the desired grasp pose. In the reflex approach, fingers remain open until the thumb makes contact with the object. In the naive approach, the fingers are treated like a parallel jaw gripper prior to closure, leading to ground collisions for small objects.}
     %\textbf{Iterative closure} takes as input a desired width and an offset approach width for pre-grasping, and steps from the approach-width parameterized pose until closure about the grasp point at the desired width. \textbf{Reflex closure} orients and positions the hand immediately at the desired-width parameterized pose, but only actuates the thumb to the planned position, keeping the other fingers fully open. When the hand has reached the grasp point with the thumb contacting or near contact with the object, the other fingers snap shut, like a reflex. \textbf{Naive closure} moves the hand, fully open, to the grasp point, and then is commanded to the width-parameterized pose. For small objects, this can result in safety-violating ground collisions.}
  \label{fig:grasp_strategies}
  \vspace{-1mm}
\end{figure}
\subsection{Width-to-Grasp Quadratic Program Optimization}

As a complementary validation, we formulate the same width-to-grasp problem as a quadratic program (QP) solved via differential inverse kinematics using the \textsc{mink} library~\cite{Zakka_Mink_Python_inverse_2026}.
Rather than deriving coplanarity and width constraints analytically, the QP planner minimizes deviations from target tip positions---the \emph{same} positions computed by the analytical method---subject to joint velocity limits, configuration limits, and finger-coupling constraints.

At each IK step, \textsc{mink} solves:
\begin{equation}
  \min_{\dot{q}}\; \tfrac{1}{2}\|\dot{q}\|^2 \quad\text{s.t.}\quad
  J\,\dot{q} = e, \quad \dot{q}_{\min} \leq \dot{q} \leq \dot{q}_{\max},
\end{equation}
where $J \in \mathbb{R}^{(3m+k)\times n}$ stacks, for each of $m$ active fingertips, the $3\times n$ Cartesian Jacobian $\partial \mathbf{p}i/\partial q$ (relating joint velocities to tip velocity), and $k$ rows encoding linearized rigid-linkage coupling (intermediate with proximal) constraints of the form $\dot{q}_{\text{int}} = b,\dot{q}_{\text{prox}}$; and $e$ collects the corresponding tip position residuals and constraint violations.
Starting from the open hand, the solver iterates until all fingertip errors fall below $5\!\times\!10^{-4}$\,m or a maximum iteration of 1000 is reached.

\textit{Comparison with analytical results.}
Both planners converge to highly similar hand configurations and identical ranges (\ref{tab:width_ranges}): the QP implicitly discovers the same width-dependent tilt that the analytical method derives explicitly, validating that the geometric constraints are consistently captured by both formulations.
On 3, 4, and 5-finger grasps, the QP achieves marginally lower Z-span, showing it can occasionally find a slightly flatter contact arrangement.

However, even when converged or at max iterations, the QP incurs a nonzero mean tip position error of 2.4--3.3\,mm depending on mode, compared to zero by construction for the analytical method (Tab. \ref{tab:width_ranges}). For precision pinch grasps on small or thin objects, this residual offset is consequential.
We therefore retain the analytical method as the primary planner and present the QP comparison as a structural validation.

\begin{figure}
  \centering
  \includegraphics[width=0.95\columnwidth]{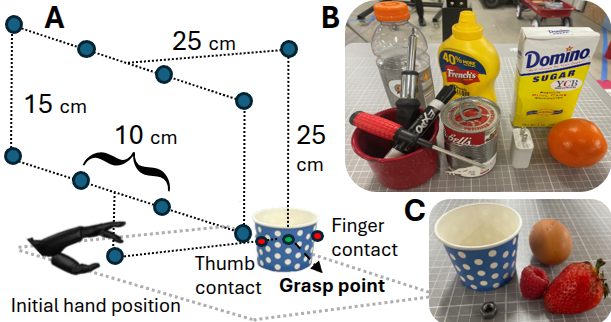}
  \caption{\textbf{A.} For all evaluated objects, we assume high-fidelity perception of the major axis and centroid for antipodal grasping. Then, we define 10 offset positions with randomized orientation from which to initiate grasping \textbf{B.} We evaluate on 10 YCB and YCB-like (bottle) objects, adapted from \cite{FromPowerToPrecision2024}. \textbf{C.} We additionally grasp 5 delicate objects to evaluate adaptive grasping.}
  \label{fig:grasp_objects}
  \vspace{-1mm}
\end{figure}

\vspace{-1mm}

\begin{figure*}[t]
  \centering
  \includegraphics[width=0.9\textwidth]{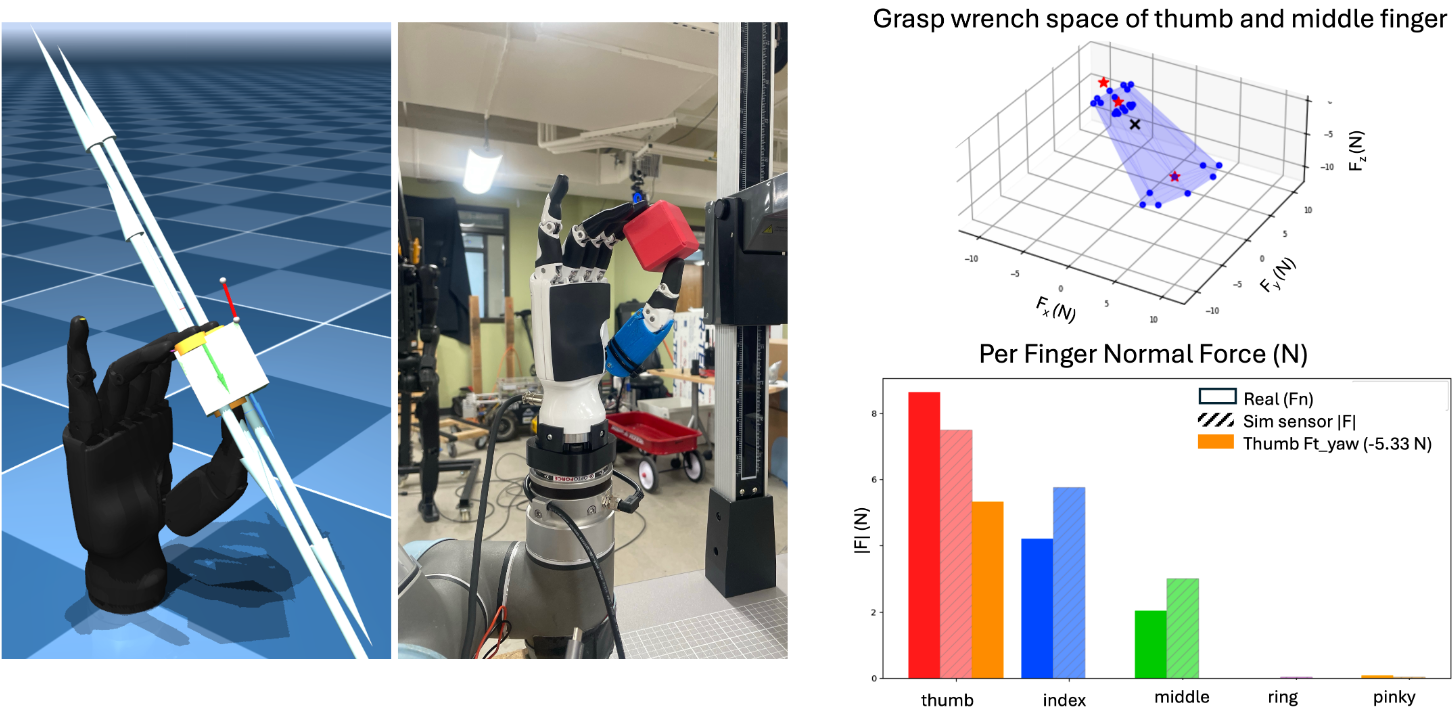}
  \vspace{-1pt}
  \caption{
  MuJoCo simulation (\textbf{left}) and real hardware (\textbf{middle}) executing a precision pinch grasp (Inspire RH56DFX). Yellow discs indicate contact points detected by the simulator; white arrows show simulated contact forces. Forces from simulation and real measurements (\textbf{bottom right}), as well as corresponding grasp wrench space (top right). 
  %Contact stability validation and sim-to-real force comparison. (A) Contact stability analysis: convex hull of the linearized friction cones (shaded region), blue dots indicating vertices of the linearized cones, and X marking the location of the Ferrari-Canny sphere, which characterizes the center and radius of maximum achievable task-space wrench. (B) Comparison of simulated (dashed lines) and measured (solid lines) per-finger normal forces in Newtons across the contact duration, showing approximately 20\% sim-to-real mismatch, validating the MuJoCo contact model's fidelity for planning and control synthesis.
  \label{fig:sim_to_real}
  \label{fig:mujoco_pinch}}
  \vspace{-1mm}
  \vspace{-1mm}
  \vspace{-1mm}
\end{figure*}

\subsection{Grasp Strategy Benchmark }
We design three grasp strategies derived from analytically planned grasps, shown in Fig.~\ref{fig:grasp_strategies}. Iterative closure takes as input a desired width and an offset approach width for pre-grasping. It initiates at the approach-width parameterized pose and steps down until closure about the grasp point at the desired width. In comparison, reflex closure positions the hand at the final width-parameterized pose while the other active fingers stay fully open, such that only when the thumb is placed against the object as a backstop do the remaining fingers close. This minimizes the window during which fingers can displace the object. Finally, naive closure moves the hand from an arbitrary pose and fully open to the grasp point, like a parallel jaw gripper. Only then does it command the width-parameterized pose in one step, making the strategy susceptibility to ground collision for small objects. For all grasps we employ a final hybrid force control policy to reach a desired contact force.

We validate planner-generated grasps on representative objects (Fig.~\ref{fig:grasp_objects}), comprising 15 items spanning a broad range of widths, shapes, and fragilities. The YCB ~\cite{Calli_2015} subset (10 objects with one YCB-like bottle) covers everyday rigid objects---tools, cans, boxes, and containers---selected to exercise the full width range of the planner and to match the evaluation set of Ye et al.~\cite{FromPowerToPrecision2024} for reference comparison. The delicate subset (raspberry, strawberry, paper cup, egg, M6 nut) includes items where excessive grip force risks damage (or pressure-induced slips for the nut), requiring the adaptive force-control policy rather than position-only closure.

The evaluation protocol assumes high-fidelity perception: object width, major axis orientation, and centroid are provided to the planner, which then computes the full hand configuration (joint angles, hand tilt, and wrist offset) analytically from geometry alone.
From each object's grasp plane, we sample 10 approach poses, including one adversarial grasp with the hand level with the grasp point Z, susceptible to object collision pre-grasp. Then we apply randomized lateral offsets ($\pm$ 3\,cm) and orientation around the approach axis, yielding 100 YCB trials and 50 delicate trials (150 total per strategy).
A trial is counted successful if the hand lifts 20\,cm vertically at 0.1\,$\frac{m}{s}$ and holds the object stably through the full grasp sequence without dropping or damaging it.

For the rigid YCB objects, we command a fixed desired force of 6\,N, distributed across the active fingers. For delicate objects, we use the same estimated forces from Xie et al. ~\cite{DeliGrasp} and verify with 100 queries on Claude Sonnet 4.6 per object that the mode of property estimates and their resulting forces are $\pm$10\% of the selected forces. We select the \textit{n}-finger grasp to cover the half-length of the object major axis.

\subsection{Contact Stability and Grasp Quality}
We provide online grasp quality analysis for manipulation. Fig.~\ref{fig:mujoco_pinch}, Left, demonstrates real2sim contact analysis on a pinch.
For each contact, the contact wrench must lie within the linearized friction cone defined by the friction coefficient $\mu$ and contact geometry, shown as yellow discs and directional arrows.
The grasp wrench space is the convex hull of all contact wrenches~\cite{FerrariCanny1992} shown in \ref{fig:sim_to_real}, Top Right.
A grasp is stable if the desired task wrench (e.g., gravitational force or applied load) lies within the convex hull of the grasp wrench space~\cite{FerrariCanny1992}. Our simulation shows an approximate 20\% sim (dashed)-to-real (solid) mismatch, Fig. \ref{fig:sim_to_real}, Bottom Right.

\par Force closure is not currently incorporated as a closed-loop control signal, and we have not yet fully characterized the sim-to-real gap across the full range of contact configurations and object geometries. In practice, the live visualization serves two roles: pre-grasp verification, when a planned grasp can be simulated and its force-closure metrics inspected to adjust commanded force, and post-grasp constraints on end-effector velocity to maintain a minimum force-closure margin or triggering re-grasping when contact quality degrades.
\vspace{-1mm}
\vspace{-1mm}
\vspace{-1mm}
\vspace{-1mm}
\vspace{-1mm}
\subsection{Open-source usage}
The described simulation, grasp planning, alternative optimization, and grasp quality analysis are packaged in configurable Tkinter GUI applications, supporting sim-only, real-only, and sim2real interaction. We provide pure Python and ROS2 interfaces for hardware, control, and planning.

%%%%%%%%%%%%%%%%%%%%%%%%%%%%%%%%%%%%%%%%%%%%%%%%%%%%%%%%%%%%%%%%%%%%%%%%%%%%%%%%
\vspace{-1mm}
\section{Experiments}
\vspace{-1mm}
\label{sec:experiments}
\subsection{Hybrid Speed-Switching for Overshoot Elimination}
\vspace{-1mm}

Because the RH56 executes contact motion with little pre-deceleration, high-speed closure produces large force overshoot.
We therefore evaluate the hybrid policy introduced above, which combines fast free-space motion with low-speed contact. 
Fig.~\ref{fig:overshoot} shows that overshoot is governed primarily by contact-phase speed: overshoot increases sharply with constant commanded speed, whereas the hybrid condition (1000+25) remains nearly identical to constant $v=25$ across all tested force setpoints.
Fig.~\ref{fig:completion_time} shows that this overshoot reduction is achieved without the large time penalty of constant low-speed motion, since only the final contact segment is executed at $v=25$.
\vspace{-1mm}
\begin{figure}[t]
  \centering
  \includegraphics[width=0.95\columnwidth]{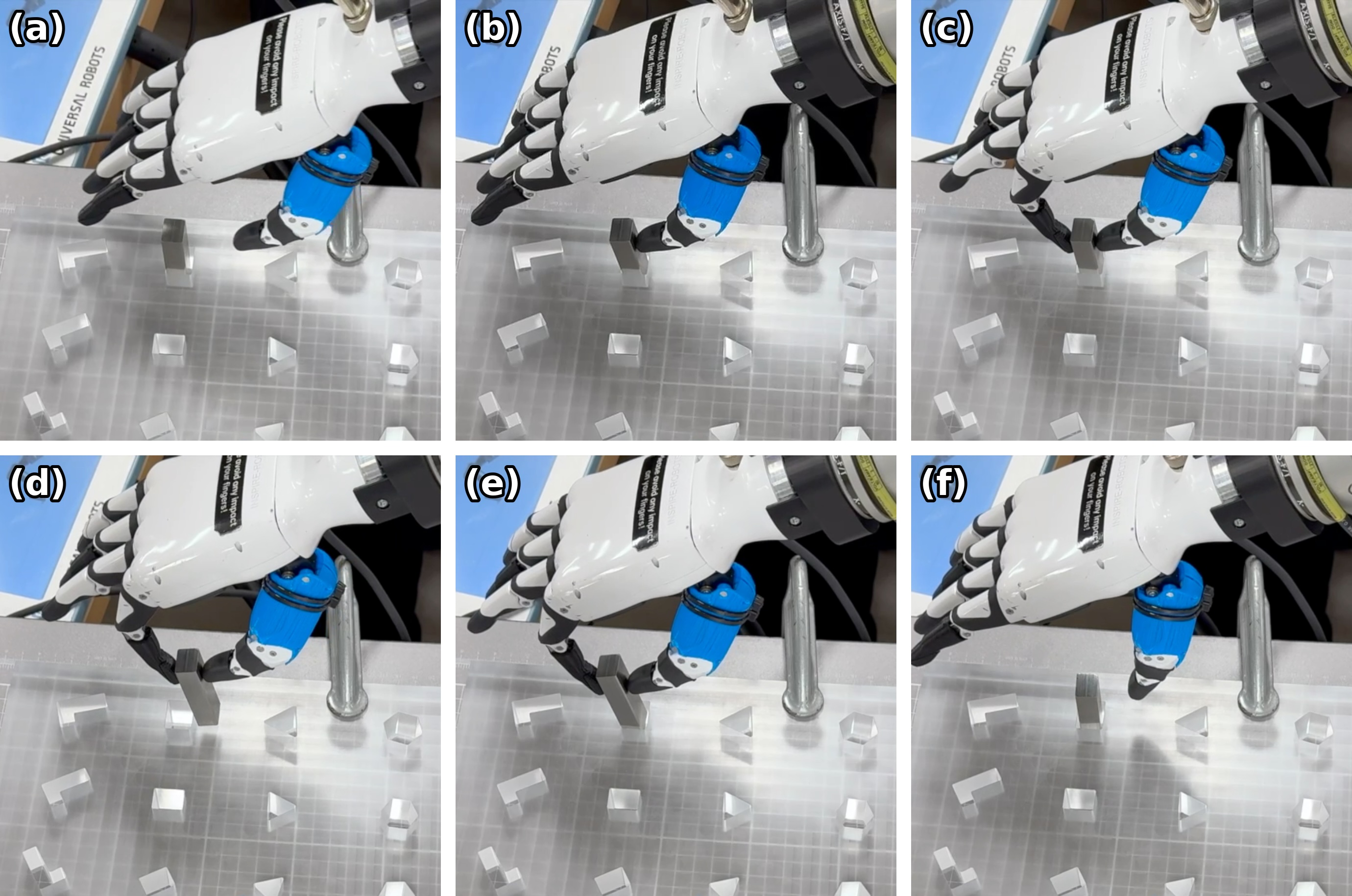}
  \vspace{-1mm}
  \vspace{-1mm}
  \caption{Reflex closure grasp and thumb force-triggered release in the manipulation-net peg-in-hole benchmark (quadratic peg). (\textbf{a--c}) The thumb applies force at the peg until triggering index closure to secure the peg (\textbf{c}). The arm moves up and away (not shown), and then in (\textbf{d--e}) slides back across the board to reinsert the peg. A brief thumb force spike indicates lateral contact with the hole wall, followed by a force drop when the arm retracts. These two events gate the release decision. (\textbf{f}) The fingers open to release the peg after the spike and drop are observed.}
  \label{fig:peginhole_sequence}
  \vspace{-1mm}
  \vspace{-1mm}
  \vspace{-1mm}
  \vspace{-1mm}
  \vspace{-1mm}
  \vspace{-1mm}
\end{figure}

\subsection{Peg-in-Hole Benchmark and Strategy Comparison}

We evaluate grasp-and-release strategies in a vision-free peg-in-hole setup using the quadratic peg from the manipulation-net benchmark, with 20 trials per strategy.
The peg is placed at a fixed, known location.
In this task, naive closing is not sufficient.
Because the peg is only partially exposed, it often fails to make contact at all.
We also find that successful lift and insertion require an antipodal grasp.
Without sufficient opposition, the peg tilts during closure, causing slip or misalignment while pushing.
Explicitly utilizing antipodal contact geometry from grasp planning therefore improves both grasp acquisition and insertion robustness.

\begin{figure}[t]
  \centering
  \resizebox{\columnwidth}{!}{\input{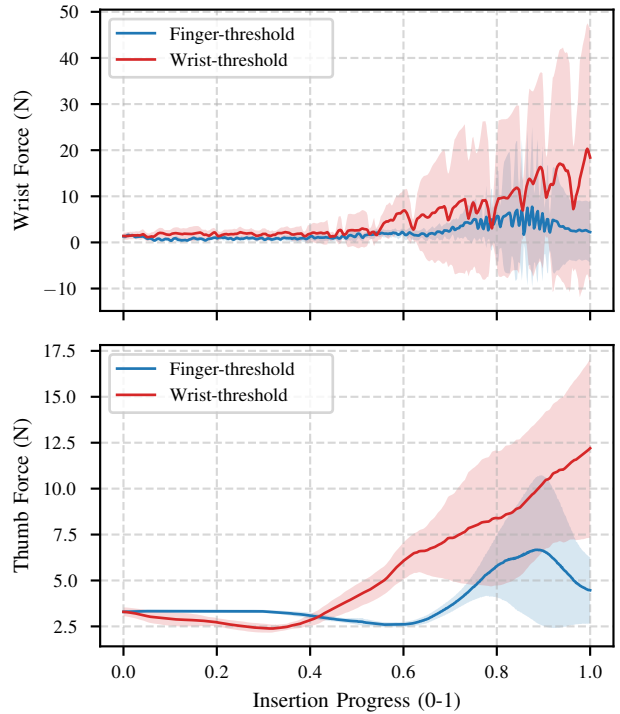}}
  \vspace{-1mm}
  \vspace{-1mm}
  \vspace{-1mm}
  \vspace{-1mm}
  \vspace{-1mm}
  \vspace{-1mm}  
  \vspace{-1mm}
  \vspace{-1mm}
  \caption{Insertion force comparison for two release policies: using finger-force threshold to trigger release (blue) and using wrist-force threshold (red). Wrist F/T (top) and thumb force (bottom). The wrist-threshold policy (red) exhibits higher force magnitudes and higher variability, as well as lower peg-in-hole success (2/20) vs. finger-threshold policy (13/20).
  \label{fig:phase2_intrinsic}} 
  \vspace{-1mm}
  \vspace{-1mm}
  \vspace{-1mm}
  \vspace{-1mm}
\end{figure}

After grasping, the arm executes a pre-programmed slow insertion trajectory toward the hole.
We compare two sensing modalities for release triggering: intrinsic finger-force sensing and wrist-force sensing.
For both modalities, release thresholds are defined relative to non-contact baseline noise.
The estimated standard deviation is approximately 0.12\,N for the finger sensor and 1.1\,N for the wrist sensor.
We evaluate both $3\sigma$ and $10\sigma$ thresholds.
The lower threshold often causes premature release during lift and insertion due to trial-to-trial contact and force fluctuations exceeding the threshold.
We therefore use the more conservative $10\sigma$ threshold in the main trials for both sensing modalities.

With this setting, finger-force release succeeds in 13/20 trials, whereas wrist-force release succeeds in 2/20 trials.
As shown in Fig.~\ref{fig:phase2_intrinsic}, the wrist-force condition produces larger contact forces and higher variability over insertion progress, which is consistent with its lower success rate.
% \subsection{Peg-in-Hole Benchmark and Strategy Comparison}

% We evaluate grasp-and-release strategies in a vision-free peg-in-hole setup using the quadratic peg from the manipulation-net benchmark (20 trials per strategy where reported).
% The peg is placed at a fixed, known location.
% Robust execution in this setting requires the proposed grasp planning rather than a naive close-to-grasp policy.
% Because the peg is only partially exposed, simply closing the fingers around the visible portion often fails to establish contact with the peg at all.
% We therefore initiate grasping with thumb reflex, where the thumb makes first contact and thumb-force reading confirms peg contact and triggers index-finger closure (Fig.~\ref{fig:peginhole_sequence}).
% Moreover, forceful lifting and subsequent pushing demand an approximately antipodal grasp.
% Without antipodal opposition, the peg readily tilts during closure, particularly due to its low-friction surface and the larger normal forces required to lift it securely.
% This initial tilt propagates into the insertion stage: even if a grasp is formed, a tilted peg is prone to slip or be pulled out of the fingers while being pushed along the board toward the hole, causing premature failures.
% By explicitly targeting antipodal contact geometry, the proposed planning improves robustness for both grasp acquisition and downstream forceful manipulation.
\begin{figure*}[t]
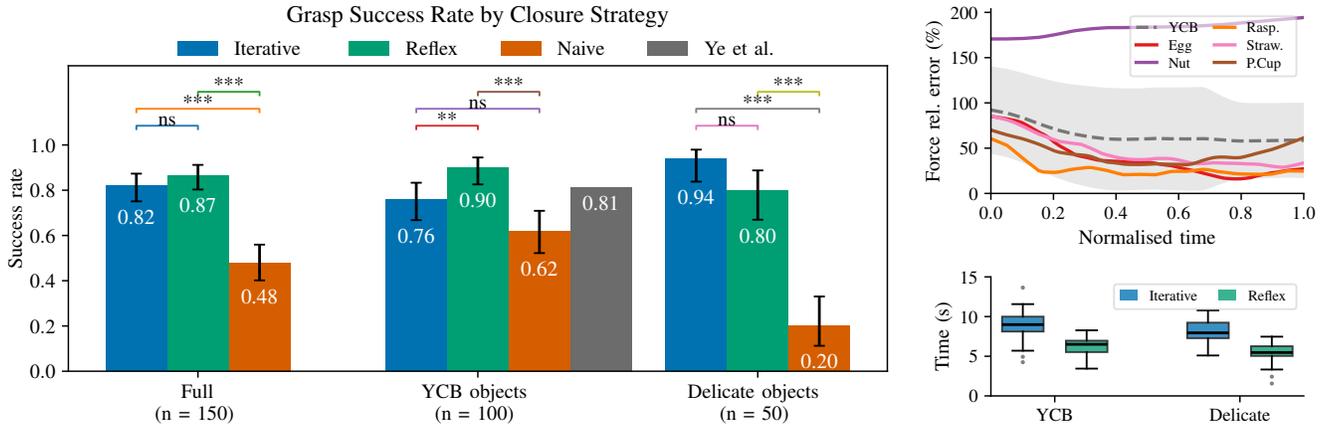

  \centering
  \resizebox{\linewidth}{!}{%
  \input{plot/grasps.pgf}\hspace{0.5em}\input{plot/side_panel.pgf}}
  \vspace{-1mm}
  \vspace{-1mm}
  \vspace{-1mm}
    \caption{
    \textbf{Left:} Grasp success rates for three closure strategies across the full experiment
    ($n=150$), YCB objects ($n=100$), and delicate objects ($n=50$).
    Bars show empirical success rates and error bars denote Wilson confidence
    intervals. Both iterative and reflex closure achieve strong performance across all object
    categories. %Pairwise statistical significance between methods is
    %evaluated using two-proportion $z$-tests with Bonferroni correction
    %($\alpha=0.0167$); 
    Significance levels are indicated above each group
    ($^{***}p<0.001$, $^{**}p<0.01$, $^{*}p<0.05$, ns: not significant).
    For the YCB subset, we include results from Ye et al. ~\cite{FromPowerToPrecision2024} on a highly similar evaluation set for
    reference. \textbf{Top Right:} Force relative error (\%) as a function of normalized force-control time. Coloured lines: individual delicate objects (per-object mean); dashed blue: YCB pooled mean $\pm 1\sigma$ band. \textbf{Bottom Right:} Total grasp time stratified by strategy: (Iterative, blue and Reflex, green) and object category: YCB and Delicate.
    }
  \label{fig:grasps}
  \vspace{-1mm}
  \vspace{-1mm}
  \vspace{-1mm}
\end{figure*}

\subsection{Grasp Strategy Benchmark Results}
We evaluated three grasp closure strategies: naive closure, reflex closure, and iterative closure.
Across the full experiment (150 trials / strategy), iterative and reflex closure achieved success rates of
$82.0\%$ and $86.7\%$, respectively, compared to $48.0\%$ for naive closure.
On the YCB object subset (100 trials), reflex closure achieved $90\%$ success compared to
$76\%$ for iterative closure and $62\%$ for naive closure, with reflex closure 
outperforming both alternatives. For the YCB subset, Fig. \ref{fig:grasps} overlays a result of 81\% success from Ye et al.~\cite{FromPowerToPrecision2024}, who evaluate a learned policy with a 12-DoF xHand (also requiring high-fidelity perception) on a highly similar object selection from the YCB benchmark. 
On delicate objects (50 trials), iterative closure achieved
the highest success rate ($94\%$), followed by reflex closure ($80\%$), both outperforming naive closure ($20\%$). Statistical significance between methods is evaluated using pairwise
two-proportion z-tests with Bonferroni correction ($\alpha = 0.0167$), with error bars denoting 95\% Wilson confidence intervals.
These results show that both reflex and iterative closure significantly outperform naive closure ($p<10^{-9}$),
while the difference between reflex and iterative closure is not statistically significant
($p=0.27$).

\emph{Qualitative Analysis:} Both iterative and reflexive closure strategies suffered with the lone adversarial grasp position, with the hand level with the grasp point Z, constituting the 13 and 8 failures for the strategies, respectively. The reflex strategy is more robust against adversarial grasps for small objects, as the fully extended fingers cleared the object during the approach without collision. Then, the right panels of Fig.~\ref{fig:grasps} provide two complementary views of performance.
The top right panel shows force relative error as a function of normalised hold time during adaptive grasping of each delicate object and then the combined YCB objects. Both strategies demonstrate similar relative overshoot on all objects (Iterative, 63.1\% and Reflex, 59.3\%). While relatively high, for delicate objects, this corresponds to approximately 60 units of raw force unit overshoot, which corresponds with the force overshoot error from Fig. \ref{fig:overshoot}. This error converges within the first portion of the hold duration across objects and maintains a stable steady state.
Then, the bottom right of Fig.~\ref{fig:grasps} shows that reflex closure is substantially faster (6.0s $\pm$ 1.3s) than iterative closure (8.6s $\pm$ 1.6s) across both categories since it replaces the multi-step width-stepping sequence with a single arm move to the final pose. However, the rapid reflex closure led to additional failures for the delicate objects, damaging or causing objects to skid off.

%%%%%%%%%%%%%%%%%%%%%%%%%%%%%%%%%%%%%%%%%%%%%%%%%%%%%%%%%%%%%%%%%%%%%%%%%%%%%%%%
\section{Discussion}
We present a unified hardware characterization, control, planning, and simulation framework for turning the Inspire RH56DFX from black-box hardware into a reproducible dexterous manipulation platform. This framework can be simply extended to additional Inspire RH56-line hands which also incorporate highly similar kinematic coupling and hand morphology, as well as to other dexterous hands altogether. Future work can include incorporating grasp quality analysis in closed-loop control, refining our antipodal grasps ~\cite{1241860} in simulation for more complex, functional grasping, and evaluating the framework with a real, imperfect perception platform. We discuss this strong assumption of high-fidelity perception for antipodal grasps and the validity of our system, in the context of modern methods, below.

Vision-language models imbued with open-world knowledge, 6D object pose detection, and generative point cloud generation are powerful tools to enable precise spatial and semantic knowledge of much of the world's objects. These technologies can enable highly effective grasp synthesis, as demonstrated in \cite{DeliGrasp, tziafas2024openworldgraspinglargevisionlanguage, chu2025graspcotintegratingphysicalproperty}, three separate methods which achieve 80\%+ parallel-jaw grasp success via connecting low-level adaptive grasping with open-world object pose detection and physical property (grasp force) estimation. 

Dexterous grasp synthesis has not leveraged these same tools, due to the initial obstacle of high-dimensional control and desire for complex, dexterous, functional grasping, which have inspired primarily learned methods for this task. In fact, several recent works evaluate learned dexterous grasping on the Inspire RH56DFX.
Table~\ref{tab:external_baselines} contextualises our results within this landscape. 

\vspace{-3mm}
% [NEW TABLE]
\begin{table}[h]
\scriptsize
\caption{Grasp Synthesis Methods for the Inspire RH56DFX}
\label{tab:external_baselines}
\centering
\renewcommand{\arraystretch}{1.00}
\begin{tabular}{@{}llr@{}}
\toprule
Method & Eval Domain / Setting / Perception Required & Success \\
\midrule
Ye et al.~\cite{FromPowerToPrecision2024}
  & Hardware co-design, YCB; \textit{sim}; \textbf{Full point cloud}  & 49\% \\
Lin et al.~\cite{UniFucGrasp2025}
  & Functional grasping (tools); \textit{real}; \textbf{Full point cloud}  & 53\% \\
Fang et al.~\cite{AnyDexGrasp2025}
  & Efficiency \& generalisation; \textit{real}; \emph{\textbf{Partial point cloud}} & 78\% \\
\midrule
Ours
  & Adaptive-force \& learning free; \textit{real}; \textbf{Antipodal grasp} & 87\% \\
\bottomrule
\end{tabular}
\end{table}
\vspace{-3mm}

These methods address different evaluation domains and, crucially, \emph{also rely on high-fidelity perception} (object point clouds). To restate: we do not claim to exceed state-of-the-art learned dexterous grasping. Rather, we suggest that our pipeline is a \emph{synergistic research tool}: it can be incorporated with learned methods by providing low-level adaptive grasping and interpretable diagnostic information (per-finger calibrated force, force-closure monitoring, sim-to-real correspondence) during and after grasp-synthesis. Or it can be paired with an object detector and a high-level reasoning agent. Either way, by characterizing and simulating each component of the Inspire RH56DFX hand---hardware calibration, antipodal grasping, hybrid speed-force control, and contact quality---we hope to provide the community with a transparent, research-ready tool for dexterous manipulation. 

%%%%%%%%%%%%%%%%%%%%%%%%%%%%%%%%%%%%%%%%%%%%%%%%%%%%%%%%%%%%%%%%%%%%%%%%%%%%%%%%
\vspace{-0.5mm}
\section*{ACKNOWLEDGMENT}
\vspace{-1mm}
The authors are supported by ARPA-E grant DE-AR0001966, "Robust Robotic Disassembly of EV Battery Packs using Open-World Vision Language Models and Symbolic Replanning." William Xie is supported by the NSF Graduate Research Fellowship. N. Correll is CEO of Realtime Manufacturing, Inc., which is working on humanoids for manufacturing applications. 
\vspace{-1mm}

\bibliographystyle{IEEEtran}
\bibliography{references}

\end{document}